\documentclass{article}

\usepackage{arxiv}
\usepackage[nocompress,noadjust]{cite}
\usepackage[colorlinks=true,
            allcolors=blue]{hyperref}
\usepackage{amsmath,amssymb,amsfonts}
\usepackage{graphicx}

\def\P{{\mathbf P}}
\def\Y{{\mathbf Y}}
\def\I{{\mathbf I}}
\def\B{{\mathbf B}}
\def\b{{\mathbf b}}

\def\p{{\mathbf p}}

\title{Weakly Supervised Image Segmentation Beyond Tight Bounding Box Annotations}

\date{} 					% Or removing it

\author{ {Juan~Wang} \\
	Horizon Med Innovation, Inc.\\
	23421 S. Pointe Dr.\\
	Laguna Hills, CA 92653 \\
	\texttt{wangjuan313@gmail.com} \\
	\And
	{Bin~Xia} \\
	Shenzhen SiBright Co., Ltd.\\
	Tinwe Industrial Park, No. 6 Liufang Rd.\\
	Shenzhen, Guangdong 518052 \\
	\texttt{b.xia@sibionics.com} \\
}

\renewcommand{\headeright}{}
\renewcommand{\undertitle}{}
\renewcommand{\shorttitle}{WSIS beyond tight BB}

\hypersetup{
pdftitle={Weakly Supervised Image Segmentation Beyond Tight Bounding Box Annotations},
pdfsubject={cs.AI, cs.CV},
pdfauthor={Juan~Wang, Bin~Xia},
pdfkeywords={Multiple instance learning, Weakly supervised image segmentation, Polar transformation, Bounding box, Deep neural networks},
}

\begin{document}
\maketitle

\begin{abstract}
Weakly supervised image segmentation approaches in the literature usually achieve high segmentation performance using tight bounding box supervision and decrease the performance greatly when supervised by loose bounding boxes. However, compared with loose bounding box, it is much more difficult to acquire tight bounding box due to its strict requirements on the precise locations of the four sides of the box. To resolve this issue, this study investigates whether it is possible to maintain good segmentation performance when loose bounding boxes are used as supervision. For this purpose, this work extends our previous parallel transformation based multiple instance learning (MIL) for tight bounding box supervision by integrating an MIL strategy based on polar transformation to assist image segmentation. The proposed polar transformation based MIL formulation works for both tight and loose bounding boxes, in which a positive bag is defined as pixels in a polar line of a bounding box with one endpoint located inside the object enclosed by the box and the other endpoint located at one of the four sides of the box. Moreover, a weighted smooth maximum approximation is introduced to incorporate the observation that pixels closer to the origin of the polar transformation are more likely to belong to the object in the box. The proposed approach was evaluated on two public datasets using dice coefficient when bounding boxes at different precision levels were considered in the experiments. The results demonstrate that the proposed approach achieves state-of-the-art performance for bounding boxes at all precision levels and is robust to mild and moderate errors in the loose bounding box annotations. The codes are available at \url{https://github.com/wangjuan313/wsis-beyond-tightBB}.
\end{abstract}

% keywords can be removed

\section{Introduction}
Image segmentation is the process of partitioning a digital image into multiple image segments such that pixels in an image segment share certain characteristics and are assigned the same category label. It has been widely studied in all kinds of applications \cite{minaee2021image} since 1965. In recent years, with the development of the deep learning in medical image analysis \cite{wang2017detecting,esteva2017dermatologist,wang2018context,wang2020simultaneous}, deep neural networks (DNNs) have been used to successfully tackle a variety of image segmentation problems in a fully-supervised manner \cite{ronneberger2015u,long2015fully,chen2018encoder}. However, it is labor-intensive and expensive to collect large-scale dataset with precise pixel-wise annotations for DNN training, thus limiting the value of the image segmentation in real applications. This is especially true in medical image analysis due to the difficulty of the segmentation problems and the difficulty in recruiting qualified annotators.
%, who are usually professional medical experts

To resolve the problem mentioned above, great interests have been made in the literature to develop weakly supervised image segmentation (WSIS), the purpose of which is to substitute costly pixel-wise annotations into cost-effective annotations as supervision for image segmentation. Several types of annotations have been investigated, including image labels \cite{ahn2019weakly,wang2020self}, points \cite{bearman2016s}, scribbles \cite{lin2016scribblesup}, and bounding boxes \cite{rajchl2016deepcut,kervadec2020bounding}. This work considers bounding boxes as supervision for image segmentation. In the literature, some efforts have been made to develop WSIS adopting bounding box supervision. For example, Rajchl \textit{et al.} \cite{rajchl2016deepcut} designed an iterative optimization approach for image segmentation, in which a neural network classifier was trained using bounding box supervision. Hsu \textit{et al.} \cite{hsu2019weakly} exploited mask R-CNN to simultaneously conduct object detection and image segmentation, in which the bounding box supervision was formulated as multiple instance learning (MIL) for image segmentation. Kervadec \textit{et al.} \cite{kervadec2020bounding} imposed a set of constraints on the network output by leveraging the tightness prior of bounding boxes as supervision for image segmentation. 

\begin{figure}[!t] 
	\centering
	\includegraphics[trim=0in 0in 0in 0in,clip,width=2in]{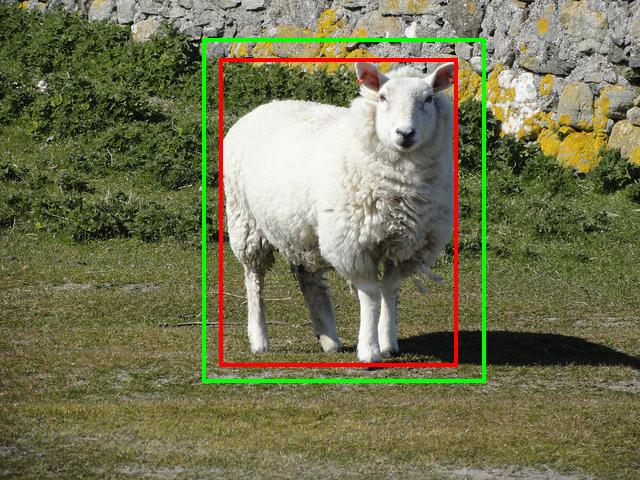}
	\caption{Demonstration of tight (red color) and loose (green color) bounding boxes for the object \textit{sheep}.}
	\label{fig:bbox_demonstration}
\end{figure}

Depending on the relationship between an object and its bounding box annotation, as shown in Figure \ref{fig:bbox_demonstration}, bounding box annotations can be divided into two categories: \textit{tight bounding box} and \textit{loose bounding box}. The tight bounding box has its four sides touching the object, thus the size of the tight bounding box is same as the size of the object; in contrast, the size of the loose bounding box is larger than the size of the object, thus at least one side of the loose bounding box does not touch the object. 
%As demonstration, the examples of a tight bounding box and a loose tight bounding box are shown in Figure \ref{fig:bbox_demonstration} for the object ``sheep''. 
Compared with the loose bounding boxes, the tight bounding boxes have strict requirements on the precise locations of the four sides of the bounding box annotations, thus it is much more difficult and time-consuming to annotate tight bounding boxes. However, most methods in the literature, if not all, achieve high segmentation performance using tight bounding box supervision, and decrease the performance greatly when supervised by loose bounding boxes \cite{kervadec2020bounding}. Such inconsistency in the difficulty of annotation acquisition and the segmentation performance for tight and loose bounding boxes poses a problem in bounding box supervision. To conquer this issue, in this paper we investigate whether it is possible to maintain good segmentation performance when loose bounding boxes are used as supervision.

Recently, in our previous study we developed a WSIS approach using tight bounding box supervision by exploiting the properties of tight bounding boxes for MIL formulation \cite{wang2021bounding} and achieved state-of-the-art segmentation performance \cite{wang2021bounding,wang2022cdrnet}. Building on our previous success in \cite{wang2021bounding}, this study extends tight bounding box supervision to loose bounding box supervision for image segmentation. For this purpose, we propose an MIL formulation based on polar transformation of the image region in the bounding box to assist the approach in \cite{wang2021bounding} for image segmentation, and develop a weighted smooth maximum approximation to incorporate the observation that pixels closer to the origin of the polar transformation are more likely to belong to the object in the bounding box. In the end, the segmentation assistance is conducted by combining the loss derived from the polar transformation based MIL into the loss in \cite{wang2021bounding}. 
The polar transformation based MIL strategy developed in this study is valid for both tight and loose bounding boxes, contributing on the good segmentation performance of the proposed approach for both tight and loose bounding box supervision. 
In the experiments, the proposed approach is evaluated by two public datasets when both tight bounding boxes and loose bounding boxes at different precision levels are used as supervision. The results demonstrate that the proposed approach outperforms several existing methods in all precision levels of bounding boxes; and more importantly, it is robust to mild and moderate errors in the loose bounding box annotations.

In summary, the contributions of this study are as follows:
\begin{enumerate}
\item First, we develop a WSIS approach beyond tight bounding box supervision. It achieves state-of-the-art performance for bounding boxes at all precision levels and is robust to mild and moderate errors in the loose bounding box annotations.
\item Second, we propose an MIL strategy based on polar transformation of the image regions of the bounding boxes to incorporate the bounding box supervision into the network output to assist the image segmentation. The proposed polar transformation based MIL formulation works for both tight and loose bounding boxes, contributing on the good segmentation performance of the proposed approach for both tight and loose bounding box supervision. 
\item Third, a weighted smooth maximum approximation is introduced for the bag prediction in the proposed polar transformation based MIL to incorporate the observation that pixels closer to the origin of the polar transformation are more likely to belong to the object in the box.
\item Finally, the proposed approach is evaluated on two public datasets when bounding boxes at different precision levels are used as supervision. The results demonstrate the effectiveness of the proposed approach for image segmentation in all precision levels of bounding boxes.
\end{enumerate}

\section{Preliminaries}
\subsection{Problem descriptions}
This study investigates the use of bounding boxes as supervision for weakly supervised image segmentation (WSIS). That is, for each object in the training set, a bounding box annotation is provided to supervise the model training.

\subsubsection{Bounding box and object} \label{section:preliminaries_bb_obj}
To avoid any ambiguity, the definitions of bounding box and object are first introduced as follows:

\textit{Bounding box} is an imaginary rectangle enclosed a thing of interest in an image, which has been widely used in object detection. For an object, its bounding box annotation encloses the whole object in the box such that it does not overlap with the region outside its bounding box. Depending on the relationship between an object and its bounding box annotation, bounding box annotations can be divided into two categories: one is tight bounding box, and the other is loose bounding box. \textit{Tight bounding box} is the smallest rectangle enclosing the whole object, thus the object must touch the four sides of its bounding box. In the end, the size of the object is same as the size of its tight bounding box. In contrary, \textit{loose bounding box} is outside of the object, thus at least one side of the loose bounding box does not touch the object. In the end, the size of the object is smaller than the size of its loose bounding box. %For demonstration, in Figure \ref{fig:bb_line_demonstration} we show examples of both tight bounding box (upper row) and loose bounding box (lower row) of the object ``sheep''.

\begin{figure}[!t] 
	\centering
	\setlength{\tabcolsep}{2pt}
	\begin{tabular}{cc}
	\includegraphics[trim=0in 0in 0in 0in,clip,width=1.5in]{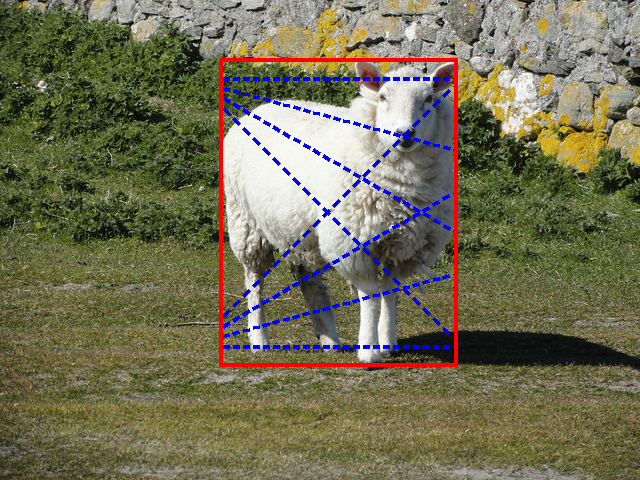} &
	\includegraphics[trim=0in 0in 0in 0in,clip,width=1.5in]{Fig2_b1.png} \\ 
	\includegraphics[trim=0in 0in 0in 0in,clip,width=1.5in]{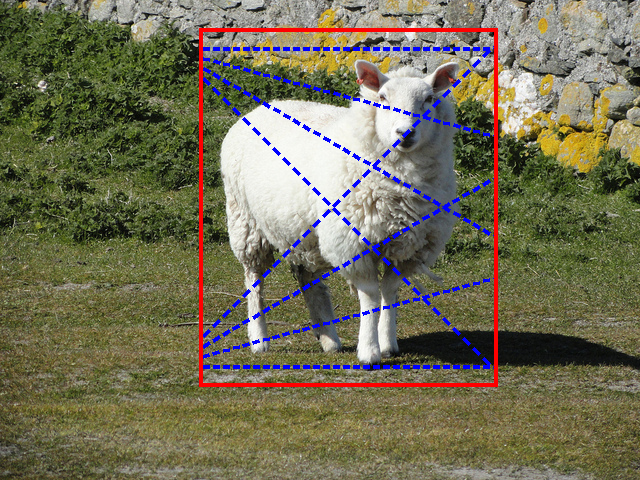} &
	\includegraphics[trim=0in 0in 0in 0in,clip,width=1.5in]{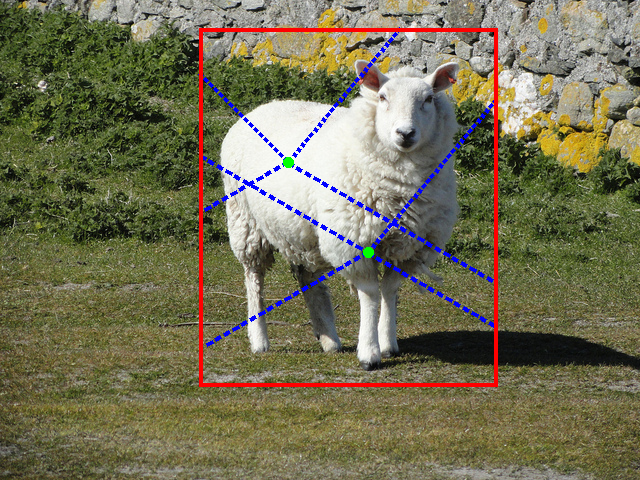} \\ 
	(a)&(b) \\
	\end{tabular}
	\caption{Demonstration of crossing and polar lines for the object ``sheep''. In these plots, crossing and polar lines are marked by blue dashed lines and bounding boxes are denoted as red rectangles, in which the tight and loose bounding boxes are shown in the upper and lower rows, respectively. The left column shows examples of crossing lines and the right column shows examples of polar lines, in which points $O$ are indicated by green dots.}
	\label{fig:bb_line_demonstration}
\end{figure}

In a bounding box, two types of lines are considered in this study. For convenience, they are named as crossing line and polar line. \textit{Crossing line} of a bounding box is defined as a line with its two endpoints located on the opposite sides of the box. \textit{Polar line} of a bounding box is defined as a line with one endpoint (denoted as point $O$) located on a pixel belonging to the object in the bounding box and the other endpoint located at one of the four sides of the bounding box. As examples, Figure \ref{fig:bb_line_demonstration} demonstrates examples of crossing lines (left column) and polar lines (right column) for both tight (upper row) and loose (lower row) bounding boxes.
%visualizes eight crossing lines (left column) and eight polar lines obtained from two different $O$'s inside the object (right column) for the bounding boxes of the object ``sheep'', in which tight and loose bounding boxes are considered for the images in the upper and lower rows, respectively. 

Without loss of generality, \textit{object} considered in this study is defined as a thing which covers a connected region in an image. It is important to notice that an object does not include any disjointed parts of a thing. 
%This definition ensures that the object in the bounding box is connected and is not disjointed. 
If there are multiple disjointed parts in a thing, each part is treated as an independent object, thus a bounding box is annotated for each part. 

\subsubsection{WSIS using bounding box supervision}
For ease of development, we first introduce the fully supervised image segmentation (FSIS). Suppose $\I$ denotes an input image and $\Y \in \{1,2,\cdots,C\}$ is its corresponding pixel-level category label for $C$ categories under consideration, the image segmentation task is to obtain the prediction of $\Y$, denoted as $\P$, for the input image $\I$. For the problem of FSIS, the pixel-wise category label $\Y$ is available for each image $\I$ in the training set for model optimization.

However, in the problem of WSIS using bounding box supervision, the pixel-level category label $\Y$ is unavailable; instead, it provides the bounding box label $\B$ as supervision for model training. In this study, the bounding box label $\B$ is denoted as $\B = \{\b_m, y_m\}, m = 1, 2, \cdots, M$, in which $M$ is the number of bounding box annotations, $\b_m$ is a 4-dimensional vector denoting the top left and bottom right points of the \textit{m}th bounding box, and $y_m \in \{1,2,\cdots,C\}$ is the category label of the object in the \textit{m}th bounding box. 

This study considers a specific type of deep neural networks (DNNs) for image segmentation, such as UNet \cite{ronneberger2015u} and FCN \cite{long2015fully}. This type of DNNs is able to output pixel-wise prediction for the input image. %That is, for each pixel in an input image, the DNN is able to determine whether it belongs to a category or not. DNNs have been widely used in image segmentation tasks and has been shown to achieve state-of-the-art performance. 
Due to the possible overlaps of objects of different categories in an image, which is especially true in medical images, this study formulates the image segmentation problem as a multi-label classification problem. That is, for a location $k$ in the input image, it outputs a vector $\p_k=[p_{k1},p_{k2},\cdots,p_{kC}]$, one element for a category; each element is converted to the range of $[0,1]$ using the sigmoid function.

\subsubsection{Multiple Instance Learning}
Multiple instance learning (MIL) is a form of weakly supervised learning in which training samples are arranged in sets, called bags, and a category label is provided for the entire bag \cite{carbonneau2018multiple}. In MIL, supervision is only provided for bags, and the labels of individual samples in the bags are not provided. For image segmentation, training samples are individual pixels of images in the training set, thus a bag consists of a set of different individual pixels of an image.

In MIL, a bag is positive if it has at least one positive sample, and a bag is negative if all of its individual samples are negative. Therefore, for a category $c$, the pixel with highest prediction in a positive bag tends to belong to category $c$, while even the pixel with highest prediction in a negative bag does not in category $c$. Based on this observation, suppose $p_{kc}$ is the network output of the \textit{k}th pixel in bag $b$ for category $c$, the bag prediction $P_c(b)$ of bag $b$ for category $c$ can be defined as 
\begin{equation}
P_c(b) = \max\limits_{k=0}^{n-1} p_{kc}
\end{equation}
where $n$ is the number of pixels in the bag $b$.

\subsection{MIL baseline} \label{section:preliminaries_mil_baseline}
This study considers the MIL baseline approach which employs \textit{tight bounding boxes} for supervision. 

\subsubsection{Positive and negative bags}
\begin{figure}[!t] 
	\centering
	\includegraphics[trim=0in 0in 0in 0in,clip,width=2in]{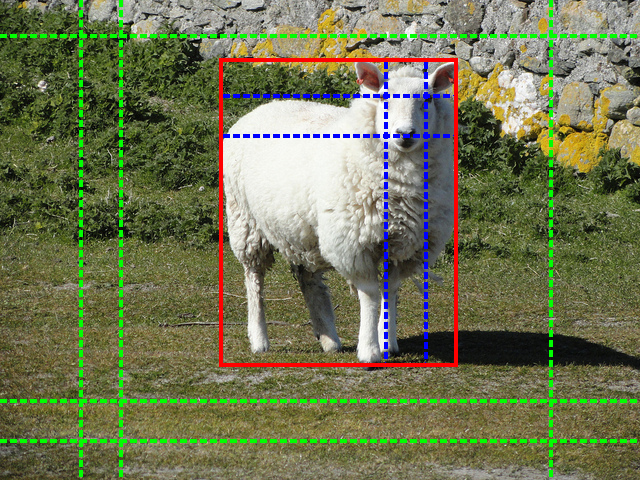} 
	\caption{Demonstration of positive and negative bags for MIL baseline with the tight bounding box annotation. The tight bounding box is indicated by the red rectangle for the object ``sheep''. The examples of positive and negative bags are marked by blue and green colors, respectively. }
	\label{fig:mil_baseline_demonstration}
\end{figure}

In an input image $\I$, for an object of category $c$ and its tight bounding box, it can be easily noted that any vertical and horizontal crossing line in the tight bounding box has at least one pixel belonging to the object in the box, hence pixels on a vertical or horizontal crossing line of the tight bounding box compose a positive bag for category $c$. Furthermore, pixels on a vertical or horizontal line of the image that do not overlap with any bounding boxes of category $c$ in the image do not belong to category $c$, hence pixels on a vertical or horizontal line of the image that does not overlap with any bounding boxes of category $c$ constitute a negative bag for category $c$. Based on these observations, for a category $c$, the MIL baseline approach considers all of the horizontal and vertical crossing lines of the tight bounding boxes of category $c$ as positive bags, and all of the horizontal and vertical lines of the image that do not overlap any bounding boxes of category $c$ in the image as negative bags \cite{hsu2019weakly}. As examples, the positive and negative bags for MIL baseline approach are demonstrated in Figure \ref{fig:mil_baseline_demonstration}.

\subsubsection{MIL baseline loss} \label{section:preliminaries_loss}
To optimize the various parameters associated with network, MIL baseline loss with two terms is employed \cite{hsu2019weakly}. For a category $c$, suppose its positive and negative bags in the training set are denoted as $\mathcal{B}_c^+$ and $\mathcal{B}_c^-$, respectively, then MIL baseline loss $\mathcal{L}_c$ is
\begin{equation}
\mathcal{L}_c = \phi_c(\P; \mathcal{B}_c^+, \mathcal{B}_c^-) + \lambda \varphi_c(\P)
\label{equ:mil_baseline_lc}
\end{equation}
where $\phi_c$ is the unary loss, $\varphi_c$ is the pairwise loss, and $\lambda$ is a constant value controlling the trade off between the unary loss and the pairwise loss.

The unary loss $\phi_c$ is defined as:
\begin{equation}
\phi_c = -\frac{1}{|\mathcal{B}_c^+|+|\mathcal{B}_c^-|} \left( \sum_{b \in \mathcal{B}_c^+} \log P_c(b) + \sum_{b \in \mathcal{B}_c^-} \log(1-P_c(b)) \right)
\end{equation}
where $|\mathcal{B}|$ is the cardinality of $\mathcal{B}$. Mathematically, the unary loss is a binary cross entropy loss for the bag prediction $P_c(b)$. It gets minimum when the the bag prediction $P_c(b)$ is 1 for positive bags and 0 for negative bags. More importantly, the unary loss adaptively selects one pixel per bag based on the network prediction for optimization, yielding an adaptive sampling effect on the training samples during training.

However, using the unary loss alone is prone to segment merely the discriminative parts of an object rather than the whole object. To resolve this problem, the pairwise loss is introduced as follows:
\begin{equation}
\varphi_c = \frac{1}{|\varepsilon|} \sum_{(k,k^\prime) \in \varepsilon} \left( p_{kc} - p_{k^\prime c} \right) ^2
\label{equ:pair_wise_term}
\end{equation}
where $\varepsilon$ is the set containing all neighboring pixel pairs. Complementary to the unary loss, the pairwise loss enforces the piece-wise smoothness in the network prediction.

In the end, for all $C$ categories, the MIL baseline loss $\mathcal{L}$ is
\begin{equation}
\mathcal{L} = \sum_{c=1}^C \mathcal{L}_c
\end{equation}

\section{Methods}
As noted in the introduction, tight bounding boxes are difficult to acquire due to the strong constraint posed on the precise locations of the four sides of the bounding box annotations. To deal with this issue, this study extends our previous approach on WSIS using tight bounding box supervision \cite{wang2021bounding}, named as parallel transformation based MIL in this study, by incorporating a polar transformation based MIL, which works for both tight and loose bounding boxes, to assist the image segmentation. In the end, the total loss $\mathcal{L}$ for network optimization is as follows:
\begin{equation}
\mathcal{L} = \sum_{c=1}^C \phi^{pa}_c(\P; \mathcal{B}_c^{pa+}, \mathcal{B}_c^{pa-}) + \phi^{po}_c(\P; \mathcal{B}_c^{po+}, \mathcal{B}_c^{po-}) + \lambda \varphi_c(\P)
\label{equ:total_loss}
\end{equation}
where $\phi^{pa}_c$ is the unary loss derived from the parallel transformation based MIL  for its positive bags $\mathcal{B}_c^{pa+}$ and negative bags $\mathcal{B}_c^{pa-}$ (will be described in Section \ref{section:method_pa_mil}), $\phi^{po}_c$ is the unary loss obtained from the polar transformation based MIL for its positive bags $\mathcal{B}_c^{po+}$ and negative bags $\mathcal{B}_c^{po-}$ (will be introduced in Section \ref{section:method_po_mil}), and $\varphi_c$ is the pairwise loss defined in equation \eqref{equ:pair_wise_term}. 

%For clarity, in this section we will first describe the parallel transformation based MIL using tight bounding box supervision, and then will give the detailed descriptions of the polar transformation based MIL using bounding box supervision. Finally, the smooth maximum approximation, a technique to deal with the issues of maximum function of bag prediction $P_c(b)$ in optimization, will be provided.

\subsection{Parallel transformation based MIL using tight bounding box supervision} \label{section:method_pa_mil}
The method described in this section was first introduced in \cite{wang2021bounding}. More details are provided in this study regarding to efficient calculation of the positive bag prediction. 

\subsubsection{Positive bags $\mathcal{B}_c^{pa+}$} \label{section:method_pa_pos_bag}
\begin{figure}[!t] 
	\centering
	\includegraphics[trim=0in 0in 0in 0in,clip,width=2in]{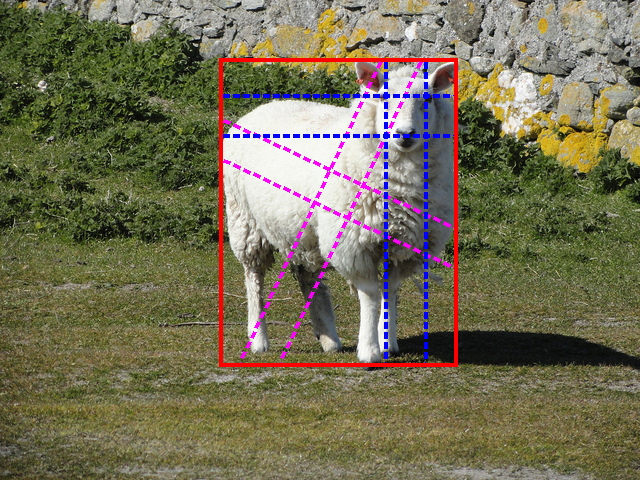}
	\caption{Demonstration of positive bags in parallel transformation based MIL with tight bounding box annotation. In this plot, the tight bounding box of the object ``sheep'' is indicated by the red rectangle; examples of positive bags from two different angles are provided, in which those with $\theta^\prime=25^\circ$ are marked by purple dashed lines and those with $\theta^\prime=0^\circ$ are given by green dashed lines.}
	\label{fig:pa_mil_demonstration}
\end{figure}

One issue associated with the positive bag definition in MIL baseline is that for an object of height $H$ pixels and width $W$ pixels, it yields only $H{+}W$ positive bags, the value of which is much smaller than the size of the object, hence limiting the selected positive samples during training and resulting in a bottleneck for the segmentation performance. Noticed that in an input image $\I$, for an object of category $c$ and its tight bounding box, any parallel crossing line of the tight bounding box also has at least one pixel belonging to the object in the box, this study generalizes the positive bag definition by considering a parallel crossing line of the tight bounding box as a positive bag for category $c$. A parallel crossing line of a bounding box can be parameterized by an angle $\theta^\prime \in (-90^\circ, 90^\circ)$ with respect to the edges of the box where its two endpoints located. For an angle $\theta^\prime$, two sets of parallel crossing lines can be obtained from the bounding box, one crosses up and bottom edges of the box, and the other crosses left and right edges. In the end, the positive bags $\mathcal{B}_c^{pa+}$ for category $c$ are all parallel crossing lines of the tight bounding boxes of the objects of category $c$ on a set of different angles. As demonstration, Figure \ref{fig:pa_mil_demonstration} shows examples of positive bags obtained from two different angles, where those indicated by purple dashed lines are with $\theta^\prime=25^\circ$, and those marked by green dashed lines have $\theta^\prime=0^\circ$. More importantly, by comparing Figures \ref{fig:mil_baseline_demonstration} and \ref{fig:pa_mil_demonstration}, the positive bags in MIL baseline are a subset and special cases of the positive bags $\mathcal{B}_c^{pa+}$ with $\theta^\prime=0^\circ$. In the experiments, this study presets the set of angles as $\theta^\prime \in (a,b,s)$, denoting evenly spaced angle values within interval $(a,b)$ with step $s$.

\begin{figure}[!t] 
	\centering
	\setlength{\tabcolsep}{2pt}
	\begin{tabular}{cc}
	\includegraphics[trim=0in 0in 0in 0in,clip,width=1.3in]{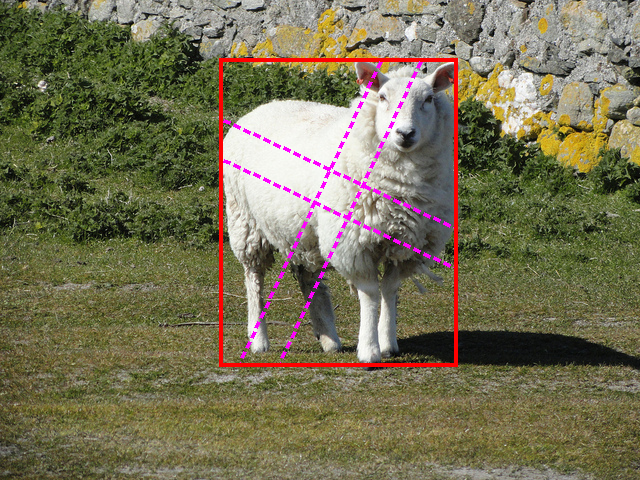} &
	\includegraphics[trim=0in 0in 0in 0in,clip,width=1.3in]{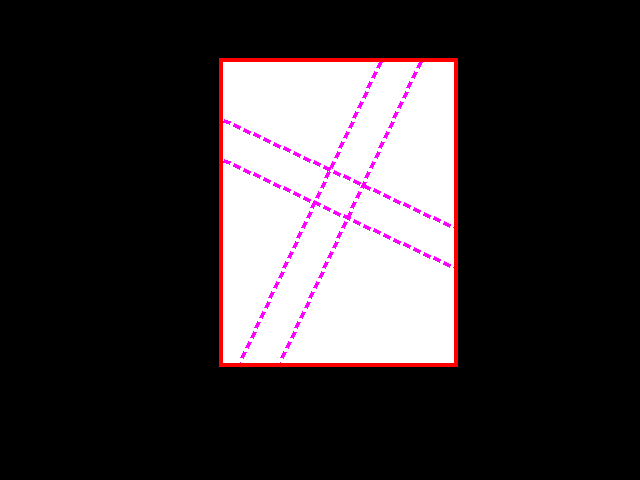} \\ 
	\includegraphics[trim=0in 0in 0in 0in,clip,width=1.6in]{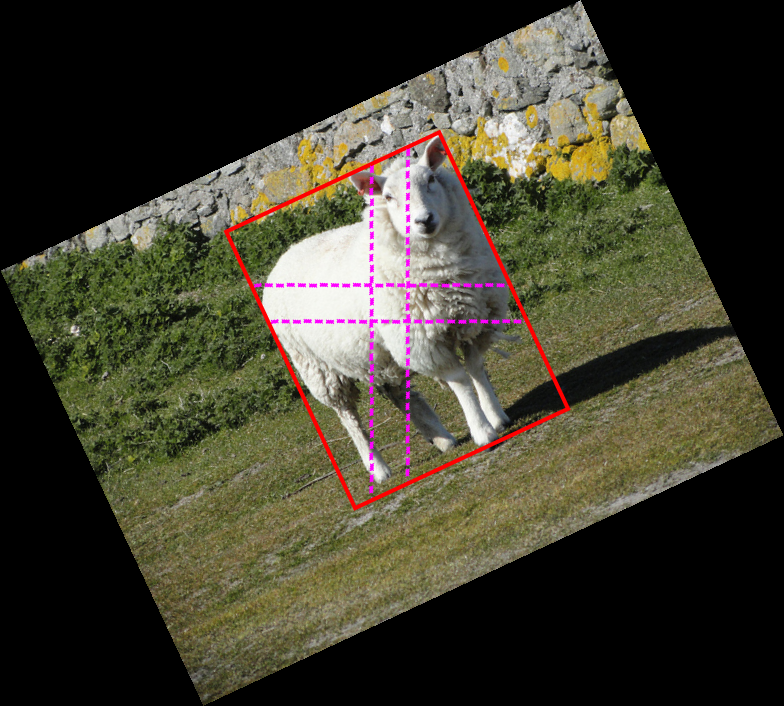} &
	\includegraphics[trim=0in 0in 0in 0in,clip,width=1.6in]{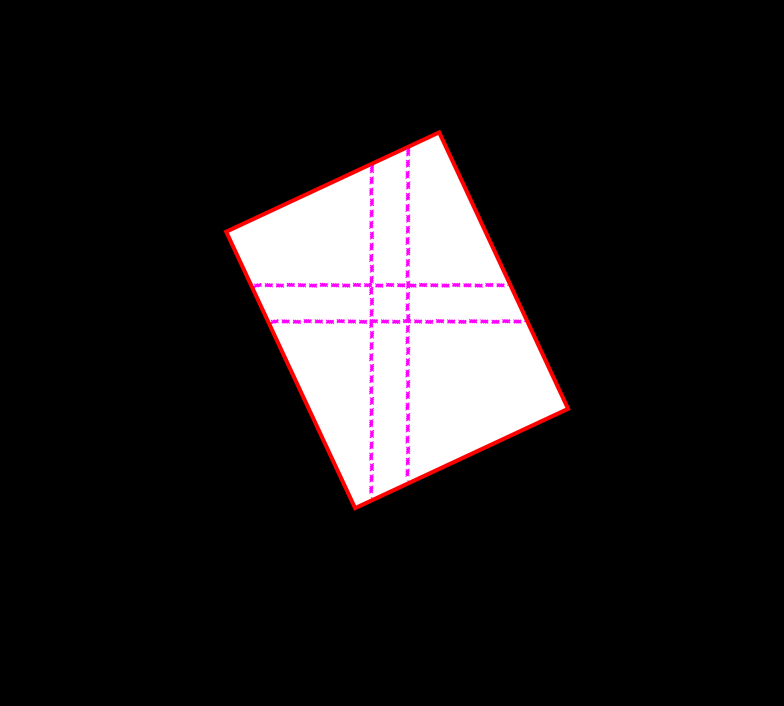} \\ 
	(a)&(b) \\
	\end{tabular}
	\caption{Demonstration of parallel transformation, in which the images and their results of parallel transformation are provided in the upper and lower rows, respectively. In these plots, examples of paralleled crossing lines with angle $\theta^\prime=25^\circ$ are marked by purple dashed lines. (a) input image, in which the tight bounding box of ``sheep'' is marked by red rectangle, (b) the box-mask image of the tight bounding box in (a).}
	\label{fig:pa_mil_transformation}
\end{figure}

In implementation, it is inefficient to directly calculate the bag prediction $P_c(b)$ of positive bags $\mathcal{B}_c^{pa+}$ from parallel crossing lines in the input image. To facilitate it, we propose to transform the parallel crossing lines with angle $\theta^\prime$ in the input image into vertical or horizontal lines by rotating the input image by angle $\theta^\prime$. In this study, this process of obtaining parallel crossing lines is named as \textit{parallel transformation} to emphasize that the transformation is targeted at crossing lines of bounding boxes. As examples, Figure \ref{fig:pa_mil_transformation}(a) shows the process of parallel transformation, in which examples of parallel crossing lines with angle $\theta^\prime=25^\circ$ in the upper image are transformed into horizontal and vertical lines in the lower image. 

However, for efficient calculation of the bag prediction, the parallel transformation of the input image alone is not enough. As shown in Figure \ref{fig:pa_mil_transformation}(a), the vertical lines in the rotated image are no longer aligned at the same starting and ending points along the horizontal direction; and the same problem also exists for the horizontal lines in the rotated images. Therefore, an indicator has to be provided for each pixel in the rotated image to determine whether it is in a positive bag or not. For this purpose, we construct a box-mask image for each tight bounding box in the input image. The box-mask image has same size as the input image, its value is set to be 1 for the pixels in the box and 0 for those outside the box. Afterwards, the same parallel transformation is applied to the box-mask image to determine whether each pixel is in a positive bag. In the rotated box-mask image, the pixels with value 1 in a vertical or horizontal line corresponds to a parallel crossing line, thus consisting of a positive bag. As examples, in Figures \ref{fig:pa_mil_transformation}(b), we also show the parallel transformation of the box-mask image of Figures \ref{fig:pa_mil_transformation}(a). From Figure \ref{fig:pa_mil_transformation}(b), the white pixels along each horizontal or vertical line (denoted by dashed blue line) consist of a positive bag in the rotated box-mask image.

Finally, to further speed up the calculation, for a tight bounding box of an object, the parallel transformation is only applied to the cropped region of the input image around the box and its corresponding box-mask image, in which a small margin is added to four sides of the box to avoid information loss during rotation.

\subsubsection{Negative bags $\mathcal{B}_c^{pa-}$} \label{section:method_pa_neg_bag}
Similar as the positive bag definition, the negative bag definition in MIL baseline also has the problem of limited samples. Notice in an input image $\I$, for a category $c$, any individual pixels outside of any bounding boxes of category $c$ in the image do not belong to category $c$, we define a negative bag for category $c$ as an individual pixel outside any bounding boxes of category $c$. In the end, negative bags $\mathcal{B}_c^{pa-}$ for category $c$ consist of all of individual pixels outside all of the bounding boxes of category $c$. This definition greatly increases the number of negative bags for training, and forces the network to learn every pixel outside bounding boxes.

\subsubsection{Unary loss $\phi^{pa}_c$} \label{section:method_pa_mil_loss}
The parallel transformed based MIL formulation above will inevitably lead to imbalance between positive and negative bags. To eliminate this issue, we borrow the concept of focal loss \cite{ross2017focal} and define the unary loss as follows:
\begin{equation}
\begin{split}
\phi^{pa}_c = -\frac{1}{N^+} \left( \sum_{b \in \mathcal{B}_c^{pa+}} \beta \left(1-P_c(b)\right)^\gamma \log P_c(b) + \right.
\\ \left.  \sum_{b \in \mathcal{B}_c^{pa-}} (1-\beta)P_c(b)^\gamma \log(1-P_c(b)) \right)
\end{split}
\label{equ:pa_uary_loss}
\end{equation}
where $N^+ = \max(1, |\mathcal{B}_c^{pa+}|)$, $\beta \in [0,1]$ is the weighting factor, and $\gamma \geq 0$ is the focusing parameter. Mathematically, the unary loss $\phi^{pa}_c$ is focal loss for bag prediction $P_c(b)$, it gets minimum when $P_c(b)$ is 1 for positive bags and 0 for negative bags.

\subsection{Polar transformation based MIL using tight or loose bounding box supervision} \label{section:method_po_mil}
This study proposes polar transformation based MIL to assist the parallel transformation based MIL for image segmentation. The proposed polar transformation based MIL works for both tight and loose bounding boxes, contributing on the good segmentation performance of the proposed approach for both tight and loose bounding box supervision. Its details are provided in this section.

\subsubsection{Positive bags $\mathcal{B}_c^{po+}$} \label{section:method_po_pos_bag}
To extend the positive bag definition beyond the tight bounding box supervision, we consider the polar line of the bounding box. For an object of category $c$, any polar line of its bounding box has at least one pixel belonging to category $c$, thus this study considers pixels in a polar line of the bounding box as a positive bag for category $c$. This definition does not employ any prior information of the bounding box, thus is valid for both tight and loose bounding boxes of the object. In the end, the positive bags $\mathcal{B}_c^{po+}$ for category $c$ are defined as all of the polar lines of the bounding boxes of the objects with category $c$. As examples, Figure \ref{fig:po_mil_demonstration} shows examples of positive bags for both tight and loose bounding boxes. 

\begin{figure}[!t] 
	\centering
	\setlength{\tabcolsep}{2pt}
	\begin{tabular}{cc}
	\includegraphics[trim=0in 0in 0in 0in,clip,width=1.5in]{Fig6_a.png} &
	\includegraphics[trim=0in 0in 0in 0in,clip,width=1.5in]{Fig6_b.png} \\ 
	(a)&(b) \\
	\end{tabular}
	\caption{Demonstration of positive bags in polar transformation based MIL for (a) tight and (b) loose bounding boxes of the object ``sheep''. In each plot, the bounding box is marked by the red rectangle, the point $O$ is denoted by the green dot, and examples of positive bags are indicated by blue dashed lines.}
	\label{fig:po_mil_demonstration}
\end{figure}

Mathematically, the bag prediction $P_c(b)$ of positive bags of an object from the corresponding polar lines can be efficiently obtained by applying polar transformation to the image. The polar transformation of an image transfers the image from the Cartesian coordinate system to the polar coordinate system, providing a pixel-wise representation in the polar coordinate system. In this study, it transfers a polar line of the bounding box of an object in an image into a horizontal line in the polar image. Suppose $(u,v)$ is the Cartesian coordinate of a pixel in the the Cartesian coordinate domain with respect to an origin $O_p$ and a radius $R_p$, and the polar coordinate is $(r, \theta)$, where $r > 0$ and $\theta \in [0, 2\pi)$ are the radial and angular coordinates, respectively. The polar transformation maps the pixel $(u,v)$ in the Cartesian coordinate plane to the corresponding pixel $(r, \theta)$ in the polar coordinate plane as follows: 
\begin{equation}
\begin{array}{c}
r = \sqrt{u^2+v^2} \\
\theta = \tan^{-1}(v/u) \\
\end{array}
\label{equ:polar_transformation}
\end{equation}
In polar transformation, the size of the polar image in the polar coordinate plane is preset by user during experiments. Suppose it is $N_r \times N_{\theta}$, where $N_r$ is the dimension of the polar axis and $N_{\theta}$ is the dimension of the angle axis.

\begin{figure}[!t] 
	\centering
	\setlength{\tabcolsep}{1pt}
	\begin{tabular}{cccc}
	\includegraphics[trim=0in 0in 0in 0in,clip,height=1in]{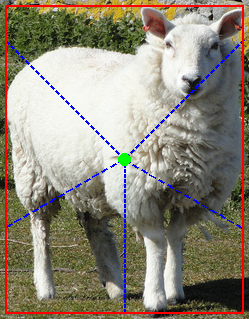} &
	\includegraphics[trim=0in 0in 0in 0in,clip,height=1in]{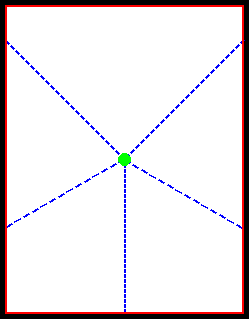} &
	\includegraphics[trim=0in 0in 0in 0in,clip,height=1in]{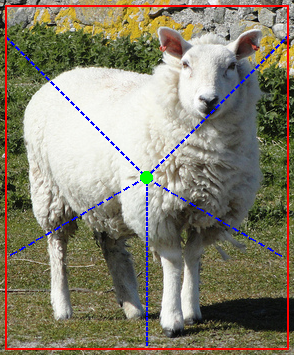} &
	\includegraphics[trim=0in 0in 0in 0in,clip,height=1in]{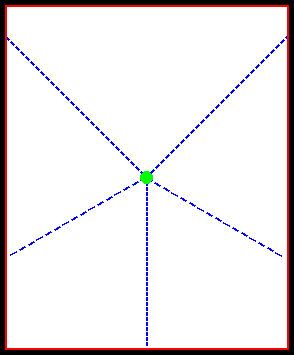} \\
	\includegraphics[trim=0.1in 0in 0in 0in,clip,height=1.2in]{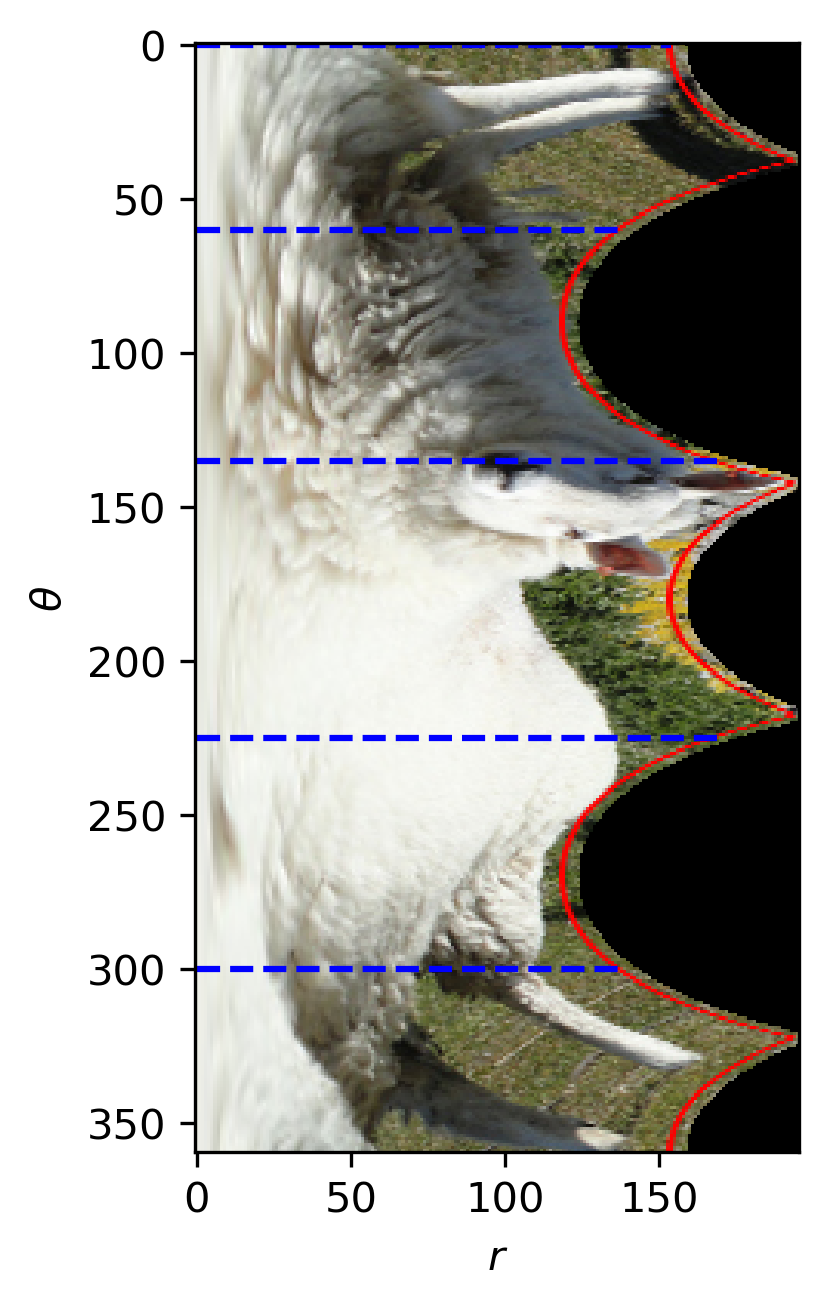} &
	\includegraphics[trim=0.1in 0in 0in 0in,clip,height=1.2in]{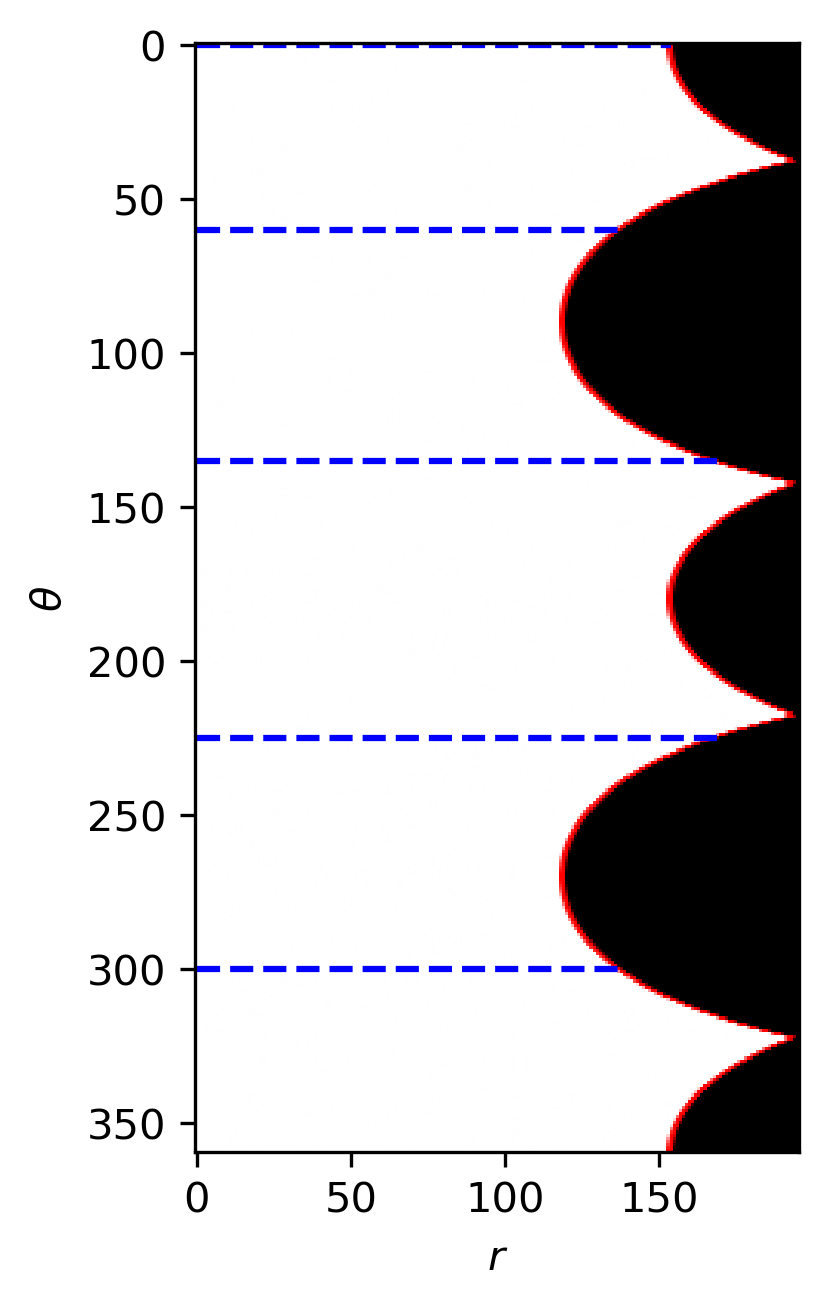} &
	\includegraphics[trim=0.1in 0in 0in 0in,clip,height=1.2in]{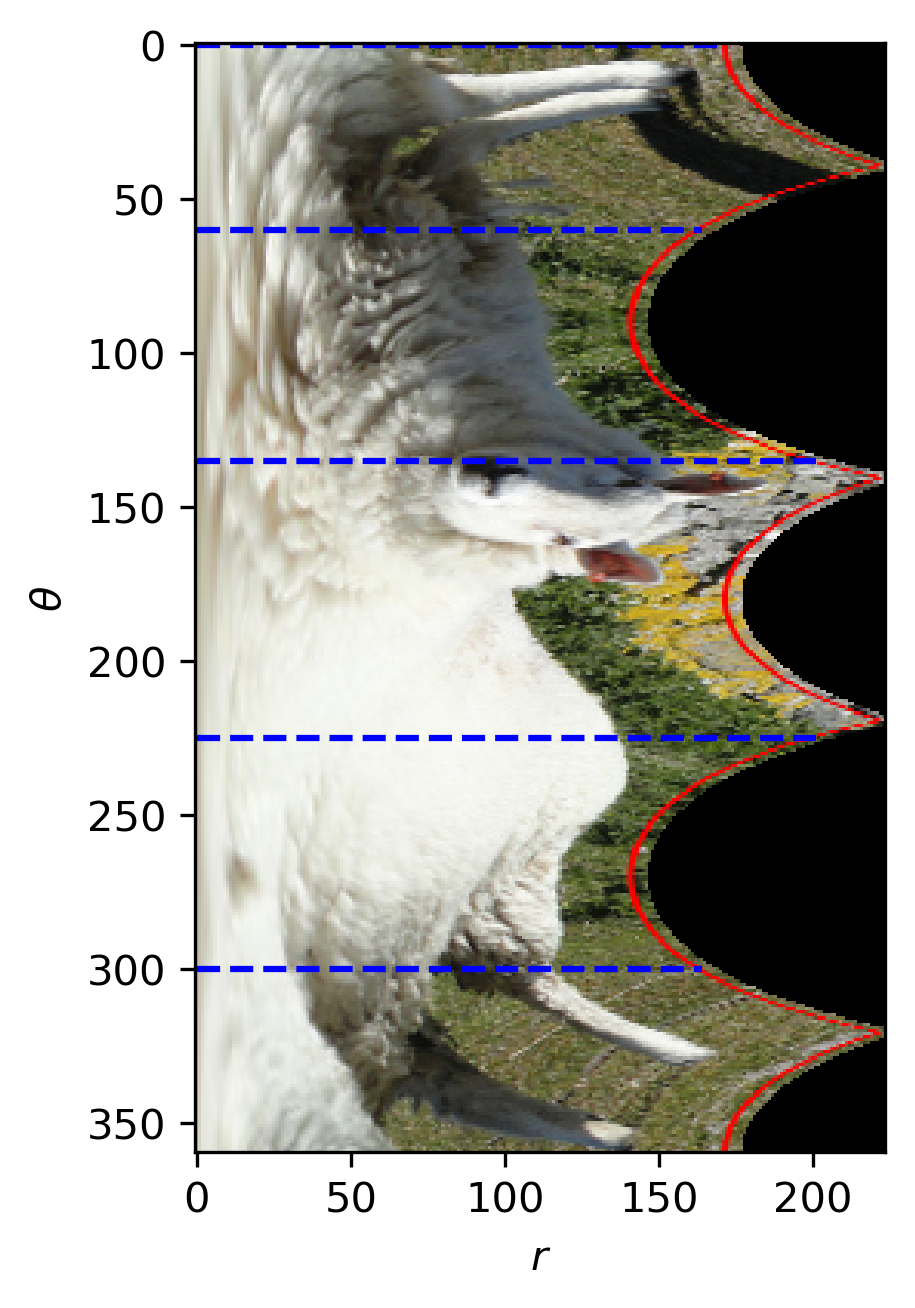} &
	\includegraphics[trim=0.1in 0in 0in 0in,clip,height=1.2in]{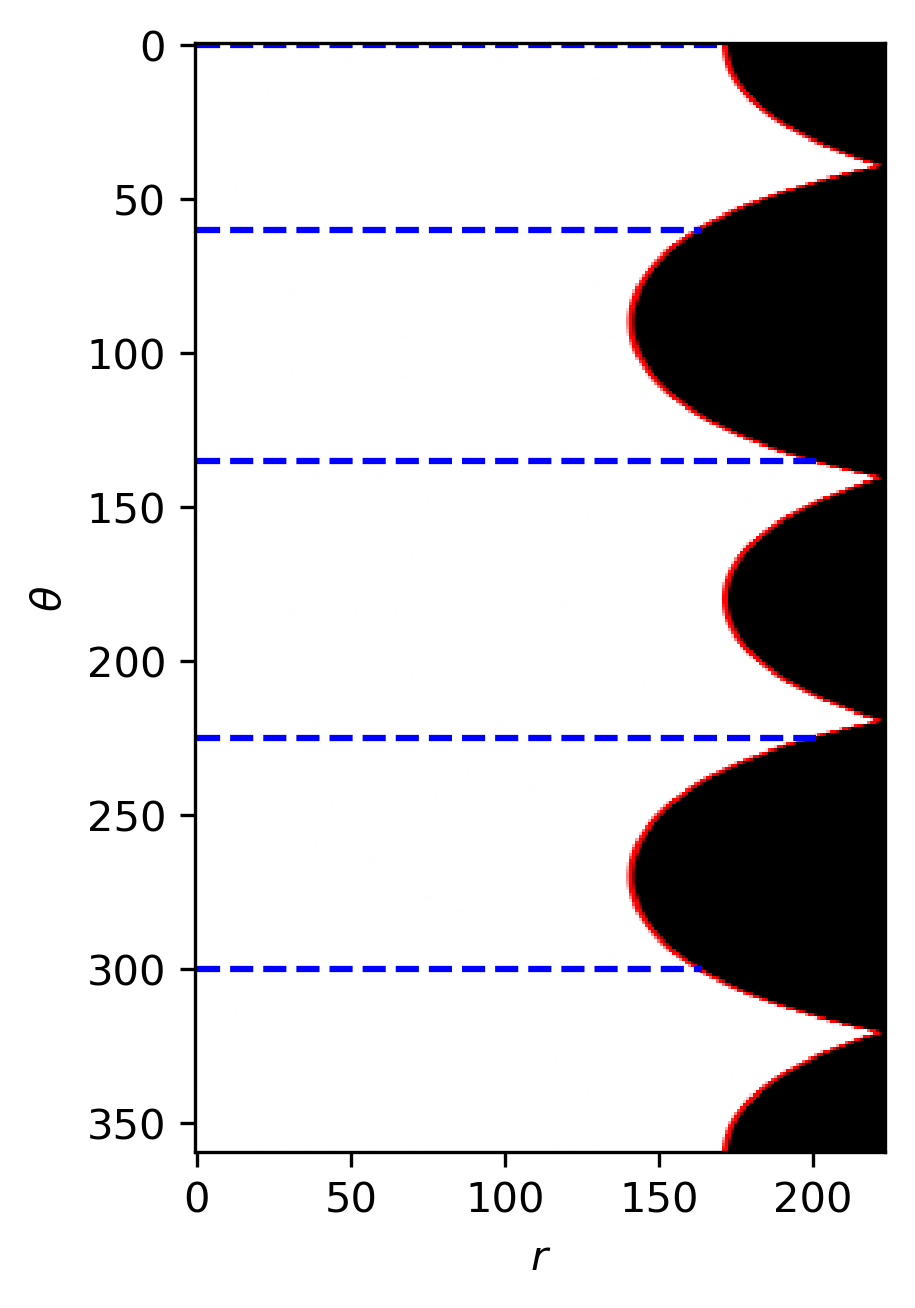} \\ 
	(a) & (b) & (c) & (d)  \\
	\end{tabular}
	\caption{Demonstration of polar transformation for examples of positive bags, in which the images and their polar images are provided in the upper and lower rows, respectively. (a) The cropped region of the object ``sheep'' from tight bounding box, (b) the box-mask image of (a), (c) the cropped region of the object ``sheep'' from a loose bounding box, (d) the box-mask image of (c).}
	\label{fig:po_mil_calculation}
\end{figure}

In this study, the polar transformation is applied to the cropped region of the input image around the bounding box. A small margin is added to four sides of the bounding box for region cropping to avoid possible information loss during transformation. For the cropped region, during polar transformation, the origin $O_p$ is set as the point $O$ defined in Section \ref{section:preliminaries_bb_obj} and the radius $R_p$ is set as half length of the diagonal line of the bounding box $R$, which is the radius of the minimum circle enclosing the bounding box. With such settings, in polar transformation of the cropped region, the radial coordinate $r$ is evenly distributed in $[0, R]$ with step $R / N_r$, and the angular coordinate $\theta$ is evenly distributed in $[0, 2\pi)$ with step $2\pi / N_{\theta}$. As demonstration, Figures \ref{fig:po_mil_calculation}(a) and (c) show polar transformation for examples of positive bags from tight and loose bounding boxes, respectively.

Similarly, as shown in Figures \ref{fig:po_mil_calculation}(a) and (c), the horizontal lines in a polar image are no longer aligned at the ending points along the vertical direction. To determine whether each pixel in the polar image is in a positive bag or not, we also introduce the box-mask image of the cropped region and applied the same polar transformation to the box-mask image as indicator of pixels in positive bags. As examples, Figures \ref{fig:po_mil_calculation}(b) and (d) show the box-mask images and its corresponding polar images for the cropped regions in Figures \ref{fig:po_mil_calculation}(a) and (c), respectively. In the lower row of Figures \ref{fig:po_mil_calculation}(b) and (d), the white pixels along a horizontal line (denoted by dashed blue lines) consist of a positive bag.

Finally, for bounding box annotation, the origin $O$ is unknown and should be determined during experiments. Based on the fact that the origin $O$ is inside the object in the bounding box, it is selected as the pixel with maximum network output among all of the pixels in the bounding box during training in the experiments. Such design is intuitive since the pixel with highest prediction are more likely belong to the object.

\subsubsection{Negative bags $\mathcal{B}_c^{po-}$} 
This study employs the same negative bag definition as in that in Section \ref{section:method_pa_neg_bag} for the polar transformation based MIL, therefore, we have $\mathcal{B}_c^{po-}=\mathcal{B}_c^{pa-}$.

\subsubsection{Unary loss $\phi^{po}_c$} 
Similar as unary loss $\phi^{pa}_c$ defined in Section \ref{section:method_pa_mil_loss}, the unary loss $\phi^{po}_c$ for polar transformation based MIL is
\begin{equation}
\begin{split}
\phi^{po}_c = -\frac{1}{N^+} \left( \sum_{b \in \mathcal{B}_c^{po+}} \beta \left(1-P_c(b)\right)^\gamma \log P_c(b) + \right. 
\\ 
\left.  \sum_{b \in \mathcal{B}_c^{po-}} (1-\beta)P_c(b)^\gamma \log(1-P_c(b)) \right)
\label{equ:po_uary_loss}
\end{split}
\end{equation}
where $N^+ = \max(1, |\mathcal{B}_c^{po+}|)$.

\subsection{Smooth maximum approximation}

In both unary losses $\phi^{pa}_c$ and $\phi^{po}_c$, the bag prediction $P_c(b) = \max\limits_{k=0}^{n-1} p_{kc}$ is the maximum prediction value of pixels in a bag. However, the derivative $\partial P_c / \partial p_{kc}$ is discontinuous, leading to numerical instability. To conquer this issue, we introduce a technique called \textit{smooth maximum approximation} to replace the maximum function by its smooth maximum approximation \cite{lange2014applications}. In this study, we consider two variants of smooth maximum approximation for $P_c(b)$ as follows:

(1) \textit{$\alpha$-softmax function:}
\begin{equation}
S_{\alpha}(b) = \frac{\sum_{k=0}^{n-1} p_{kc} e^{\alpha p_{kc}}}{\sum_{k=0}^{n-1} e^{\alpha p_{kc}}}
\label{equ:softmax_approx}
\end{equation}
where $\alpha>0$ is a constant. The higher $\alpha$ value denotes the closer approximation of $S_{\alpha}(b)$ to $P_c(b)$. %For $\alpha \rightarrow 0$, a soft approximation of the mean function is obtained.

(2) \textit{$\alpha$-quasimax function:}
\begin{equation}
Q_{\alpha}(b) = \frac{1}{\alpha} \log \left(\sum_{k=0}^{n-1} e^{\alpha p_{kc}}\right) - \frac{\log n}{\alpha}
\label{equ:quasimax_approx}
\end{equation}
where $\alpha>0$ is a constant. The higher $\alpha$ value also denotes closer the approximation of $Q_{\alpha}(b)$ to $P_c(b)$. It can be easily proved that $Q_{\alpha}(b) \leq P_c(b)$ always holds.

For image segmentation problem, the smooth maximum approximation has an extra advantage as follows: different from the maximum function $P_c$ with $\partial P_c / \partial p_{kc} > 0$ at only one pixel, the smooth maximum approximation has $\partial S_{\alpha} / \partial p_{kc} > 0$ and $\partial Q_{\alpha} / \partial p_{kc} > 0$ for all $p_{kc}$, thus it is able to learn all pixels together in a bag for model optimization. Moreover, a positive bag usually has more than one positive pixel in real segmentation problem, thus this property is beneficial as well considering this fact. Therefore, besides the advantage in numerical stability, the smooth maximum approximation is also helpful for performance improvement.

\subsubsection{Weighted smooth maximum approximation for $\mathcal{B}_c^{po+}$}
For the pixels in a polar line, the origin $O$ is inside the object and the pixels closer to the origin $O$ are more likely belonging to the object. To incorporate this fact in optimization, a weight is introduced to the smooth maximum approximation for positive bags $\mathcal{B}_c^{po+}$. In particular, a weight $w_k$ is assigned to prediction $p_{kc}$ of each pixel in the positive bag, yielding weighted smooth maximum approximation. The weight $w_k$ is defined as follows:
\begin{equation}
w_k = e^{-k^2/(2\sigma^2)}
\end{equation}
where $\sigma = (N_r-1)/\sqrt{-2\log w_{min}}$ and $w_{min}$ is a preset parameter for the minimum weight of the pixel in positive bags of an object. %Among the pixels in a polar line, the minimum weight is given to the pixel furthest from the origin $O$, which is the pixel in one of four sides of the bounding box.

\section{Experiments}

\subsection{Datasets}
This study made use of two public medical datasets as follows for performance evaluation: one is PROMISE12 dataset \cite{litjens2014evaluation} for prostate segmentation, and the other is ATLAS dataset \cite{liew2018large} for brain lesion segmentation. 

\textit{PROMISE12:} The PROMISE12 dataset was released in MICCAI 2012 grand challenge \cite{litjens2014evaluation}. It consists of transversal T2-weighted MR images and their pixel-wise annotations from 50 patients, including both benign and prostate cancer cases. The MR images were acquired at different centers with multiple MRI vendors and different scanning protocols. The dataset was divided into two non-overlapping subsets, one subset with 40 patients for training and the other with 10 patients for validation. 

\textit{ATLAS:} The ATLAS dataset was developed by University of Southern California \cite{liew2018large}. It consists of 229 T1-weighted MR images and their pixel-wise annotations from 220 patients, acquired from different cohorts and different scanners. %The pixel-wise annotations were provided by a group of 11 experts. 
The dataset was divided into two non-overlapping subsets, one subset with 203 images from 195 patients for training and the other with 26 images from 25 patients for validation. 

For fairness of comparison, for both datasets, the images in the training and validation subsets are exactly same as those in studies \cite{kervadec2020bounding} and \cite{wang2021bounding}. Moreover, same as studies \cite{kervadec2020bounding} and \cite{wang2021bounding}, this study reports the segmentation performance for the validation subset in the results.

\subsection{Performance evaluation}
This study employs dice coefficient to evaluate the performance of the proposed approach for image segmentation. The dice coefficient has been widely used as a standard performance metric in medical image segmentation. 
%To evaluate the performance of the proposed approach for image segmentation, the dice coefficient is considered, which has been widely used in medical image segmentation. 
%Mathematically, the dice coefficient is defined as:
%\begin{equation}
%dice = 2TP / (2TP + FP + FN)
%\end{equation}
%where $TP$, $FP$, and $FN$ are true positive, false positive, and false negative, respectively. 
It is in the range of $[0, 1]$. The higher dice coefficient represents better segmentation performance. In this study, the dice coefficient is calculated based on 3D MR images by stacking predictions of the corresponding 2D slices together. 

\subsection{Bounding box settings}
To evaluate the performance of the proposed approach supervised by bounding boxes at different precision levels, this study considers both tight and loose bounding boxes in the experiments. The loose bounding box of an object is obtained by adding a margin (denoted as $m$) on each side of its corresponding tight bounding box. Specifically, the bounding boxes at the following four different precision levels are investigated: 1) tight bounding boxes (denoted as $m{=}0$), 2) loose bounding boxes obtained by adding 5 pixels to each side of the corresponding tight bounding boxes (denoted as $m{=}5$), 3) loose bounding boxes which add 10 pixels to each side of the corresponding tight bonding boxes (denoted as $m{=}10$), and 4) loose bounding boxes acquired by adding random number of pixels, generated from uniform distribution in the range of $[0, 10]$, to each side of the corresponding tight bounding boxes (denoted as $m {\sim} U(0,10)$). In the experiments, $m{=}0$, $m{=}5$, and $m{=}10$ are used to quantitatively investigate the effect of precision levels of the bounding boxes on the segmentation performance, and $m {\sim} U(0,10)$ stimulates bounding boxes acquired in real annotation task, in which the margins provided by annotators are usually different and random among different objects.

To measure the precision of a bounding box annotation, mean absolute relative difference (MARD) is introduced. This study defines MARD as the average of the absolute errors in height and width between the object and its bounding box as follows:
\begin{equation}
MARD = \frac{1}{2} \times \left( \frac{m_{x_1}+m_{x_2}}{w} + \frac{m_{y_1}+m_{y_2}}{h} \right)
\end{equation}
where $w$ and $h$ are width and height of the object, respectively; $m_{x_1}$, $m_{x_2}$, $m_{y_1}$, and $m_{y_2}$ are the margin added to the left, right, up, and down sides of the tight bounding box. %MARD is a relative precision measure of the bounding box with respect to the size of the object.

In Table \ref{table:datasets_mard}, the mean and standard deviation of the MARD values for the bounding boxes at four different precision levels are provided for PROMISE12 dataset, in which the results are calculated based on all of the bounding boxes in the training subset. As can be seen, mean MARD values are close to 22\% for $m{=}5$ and $m {\sim} U(0, 10)$, indicating that the bounding boxes are accurate and there is only mild error in the bounding box annotations. For $m{=}10$, mean MARD value increases into 44.56\%, indicating the sizes of the bounding boxes are almost 1.5 times of the sizes of the objects on average, a moderate error in the bounding box annotations.

\renewcommand\arraystretch{1.3}
\begin{table*}[!t]
\caption{Mean and standard deviation (in bracket) of the MARD values for the bounding boxes at four different precision levels for PROMISE12 and ATLAS datasets.}
\centering
\setlength{\tabcolsep}{5pt}
\begin{tabular}{ccc}
\hline
\hline
Bounding box settings & PROMISE12 & ATLAS \\
\hline
$m{=}0$ & 0.00\% (0.00\%) & 0.00\% (0.00\%) \\
$m{=}5$ & 22.28\% (10.62\%) & 122.22\% (96.68\%) \\
$m{=}10$ & 44.56\% (21.23\%) & 244.44\% (193.36\%) \\
$m {\sim} U(0, 10)$ & 22.23\% (12.90\%) & 122.67\% (109.41\%) \\
\hline
\hline
\end{tabular}
\label{table:datasets_mard}
\end{table*}

Table \ref{table:datasets_mard} also lists the mean and standard deviation of the MARD values for ATLAS dataset. It can be seen, for $m{=}5$ and $m {\sim} U(0, 10)$, the mean MARD values are close to 122.5\%, suggesting a severe error in bounding box annotations.
%that the bounding boxes are much larger than the objects and the error in bounding box annotations is severe. 
Lastly, for $m{=}10$, the mean MARD value is 244.44\%. It represents that the sizes of the bounding boxes are almost 2.5 times larger than the sizes of the objects on average, indicating a very severe error in bounding box annotations.

\subsection{Methods for comparison}
To demonstrate the overall performance of the proposed approach for image segmentation, this study considers the following methods for comparison:

1) Fully supervised image segmentation (FSIS): FSIS employs pixel-wise annotations as supervision for image segmentation. It can be treated as the upper bound of the segmentation performance for WSIS due to the use of fully supervised learning based on costly pixel-wise annotations.

2) MIL baseline: It is a WSIS approach supervised by \textit{tight bounding boxes}. It is described in Section \ref{section:preliminaries_mil_baseline}. 

3) Deep cut \cite{rajchl2016deepcut}: It is a WSIS approach using \textit{bounding box} supervision for image segmentation. It trains neural network classifier in an iterative optimization way.

4) Global constraint \cite{kervadec2020bounding}: It is a WSIS approach adopting \textit{tight bounding boxes} as supervision. It imposes a set of constraints on the network outputs based on the tightness prior of bounding boxes for image segmentation. 

5) Parallel transformation based MIL (denoted as PA): It is a WSIS approach with \textit{tight bounding box} supervision, which was developed in \cite{wang2021bounding}. This study also describes it in Section \ref{section:method_pa_mil} for completeness. The loss of this approach is $\mathcal{L} = \sum_{c=1}^C \phi^{pa}_c(\P; \mathcal{B}_c^{pa+}, \mathcal{B}_c^{pa-}) + \lambda \varphi_c(\P)$. Besides being an existing method for comparison, this approach also serves as an ablation study of the proposed approach which removes the component of polar transformation based MIL.

6) Polar transformation based MIL (denoted as PO): It is a WSIS approach supervised by \textit{bounding boxes}. It is described in Section \ref{section:method_po_mil}, optimized by the loss $\mathcal{L} = \sum_{c=1}^C \phi^{po}_c(\P; \mathcal{B}_c^{po+}, \mathcal{B}_c^{po-}) + \lambda \varphi_c(\P)$ during training. This method is an ablation study of the proposed approach after removing the component of parallel transformation based MIL.

Overall, the summary of the methods for comparison is listed in Table \ref{table:methods_for_comparison}. For fairness of comparison, the network structures used in a comparison study are same for all methods.

\renewcommand\arraystretch{1.3}
\begin{table*}[!t]
\caption{Summary of the methods for comparison.}
\centering
\begin{tabular}{ccc}
\hline
\hline
Methods & Supervision & Properties of supervision \\
\hline
FSIS & masks & pixel-wise \\
Deep cut \cite{rajchl2016deepcut} & bounding boxes & tight/loose \\
Global constraint \cite{kervadec2020bounding} & bounding boxes & tight \\
MIL baseline & bounding boxes & tight \\
PA \cite{wang2021bounding} & bounding boxes & tight \\
PO & bounding boxes & tight/loose \\
Proposed approach & bounding boxes & tight/loose \\
\hline
\hline
\end{tabular}
\label{table:methods_for_comparison}
\end{table*}

\subsection{Implementation details}
\subsubsection{Experimental setups}
In this study, all experiments were implemented using PyTorch and the experimental codes are available at \url{https://github.com/wangjuan313/wsis-beyond-tightBB}. Image segmentation was conducted on the 2D slices of MR images. As indicated below, most experimental setups were set to be same as those in \cite{kervadec2020bounding} and \cite{wang2021bounding} for fairness of comparison. 

For the PROMISE12 dataset, all images were resized to $256\times256$ pixels. A residual version of UNet \cite{ronneberger2015u} was employed for segmentation. The models were trained by Adam optimizer \cite{kingma2014adam} with parameters as follows: batch size = 16, initial learning rate = $10^{-4}$, $\beta_1 = 0.9$, and $\beta_2 = 0.99$. An off-line data augmentation procedure was performed to the images in the training set, and the following operations were considered: 1) mirroring, 2) flipping, and 3) rotation.

For the ATLAS dataset, all images were resized to $208\times256$ pixels. ENet \cite{paszke2016enet} was used as backbone for image segmentation. The models were trained by Adam optimizer with following parameters: batch size = 80, initial learning rate = $5 \times 10^{-4}$, $\beta_1 = 0.9$, and $\beta_2 = 0.99$. No augmentation was conducted during training.

\subsubsection{Hyperparameters}
The weight $\lambda$ of the pairwise loss $\varphi_c(\P)$ was set as $\lambda=10$ based on experience in all experiments. The parameters in the unary losses $\phi^{pa}_c$ and $\phi^{po}_c$ were set as $\beta=0.25$ and $\gamma=2$ according to the focal loss \cite{ross2017focal}. For parallel transformation based MIL, the parameters $\theta^\prime$ for parallel crossing lines and $\alpha$ for smooth maximum approximation were obtained by grid search with the following values: $\alpha \in \{4,6,8\}$ and $\theta^\prime \in \{(-40^\circ,40^\circ,10^\circ), (-40^\circ,40^\circ,20^\circ), (-60^\circ,60^\circ,30^\circ)\}$. For polar transformation based MIL, the parameters $N_r$ and $N_{\theta}$ in polar transformation and $w_{min}$ and $\alpha$ in weighted smooth maximum approximation were obtained by grid search as follows: $N_r \in \{10,20,30,40\}$, $N_{\theta} \in \{60,90,120\}$, $w_{min} \in \{0.2,0.3,0.4,0.5,0.6,0.7,0.8\}$, and $\alpha \in \{0.5,1,2\}$. 

\section{Results}
\subsection{Performance comparison for PROMISE12 dataset}
\renewcommand\arraystretch{1.3}
\begin{table*}[!t]
\caption{Comparison of dice coefficients among different methods for the PROMISE12 dataset when bounding boxes at different precision levels are considered, in which the standard deviation of dice coefficients among different MR images is reported in the bracket. NA denotes that the result is not applicable and the symbol ``-'' indicates unavailable result.}
\centering
\setlength{\tabcolsep}{5pt}
\begin{tabular}{ccccc}
\hline
\hline
Methods & $m=0$ & $m=5$ & $m=10$ & $m \sim U(0,10)$ \\
\hline
FSIS & 0.894 (0.021) & NA & NA & NA \\
Deep cut \cite{rajchl2016deepcut} & 0.827 (0.085) & - & 0.684 (0.069) & - \\
Global constraint \cite{kervadec2020bounding} & 0.835 (0.032) & - & 0.778 (0.047) & - \\
MIL baseline & 0.859 (0.038) & 0.840 (0.046) & 0.795 (0.023) & 0.832 (0.047) \\
PA ($\alpha$-softmax) \cite{wang2021bounding} & 0.878 (0.031) & 0.868 (0.033) & 0.862 (0.041) &0.869 (0.043) \\
PA ($\alpha$-quasimax) \cite{wang2021bounding} & 0.880 (0.024) & 0.871 (0.030) & 0.856 (0.039) & 0.875 (0.031) \\
PO ($\alpha$-softmax) & 0.867 (0.022) & 0.858 (0.034) & 0.843 (0.035) & 0.858 (0.032) \\
PO ($\alpha$-quasimax) & 0.871 (0.019) & 0.859 (0.030) & 0.841 (0.038) & 0.860 (0.021) \\
Proposed approach ($\alpha$-softmax) & 0.887 (0.027) & 0.882 (0.023) & 0.875 (0.034) & 0.874 (0.026) \\
Proposed approach ($\alpha$-quasimax) & 0.887 (0.017) & 0.880 (0.029) & 0.869 (0.026) & 0.876 (0.033) \\
\hline
\hline
\end{tabular}
\label{table:promise_perf_comparison}
\end{table*}

Table \ref{table:promise_perf_comparison} gives dice coefficients of the proposed approach supervised by bounding boxes at different precision levels for PROMISE12 dataset. Two models are considered for each level, one employing $\alpha$-softmax function and the other adopting $\alpha$-quasimax function. As can be seen, the proposed approach has only minor decrease in dice coefficients for $m{=}5$, $m{=}10$, and $m {\sim} U(0,10)$ when compared with $m{=}0$, indicating that the proposed approach is robust to minor and moderate errors in bounding box annotations. %Specifically, the proposed approach gets dice coefficient of 0.882 for $m=5$ (0.56\% reduction in performance), 0.875 for $m=10$ (1.35\% reduction in performance) and 0.874 for $m \sim U(0,10)$ (1.47\% reduction in performance) for $\alpha$-softmax function.

For comparison, we also report results of the MIL baseline in Table \ref{table:promise_perf_comparison}. 
It gets dice coefficient of 0.859 for $m{=}0$, 0.840 for $m{=}5$, 0.795 for $m{=}10$, and 0.832 for $m {\sim} U(0,10)$. 
These values are much lower than their counterparts of the proposed approach for all precision levels. 

Moreover, the results of PA and PO approaches are listed in Table \ref{table:promise_perf_comparison} as well. For both approaches, two models are considered for each precision level, one using $\alpha$-softmax function and the other adopting $\alpha$-quasimax function. As can be seen, both approaches have lower dice coefficients when compared with the proposed approach at different precision levels. More importantly, comparing with both of these two approaches, the proposed approach has greater performance improvements when $m$ increases from $m{=}0$ to $m{=}5$ and $m{=}10$. These results suggest that both parallel transformation based MIL and polar transformation based MIL are effective in the proposed approach, and they are especially helpful when the error in the bounding box annotations is larger.

Furthermore, Table \ref{table:promise_perf_comparison} also provides the results of Deep cut and Global constraint approaches, which are cited from study \cite{kervadec2020bounding}. The results demonstrate that the proposed approach outperforms these two methods at a large margin for both $m{=}0$ and $m{=}10$.

Lastly, FSIS gets dice coefficient of 0.894, the upper bound of performance for WSIS. As can be seen, the proposed approach achieves performance close to FSIS for $m{=}0$ and $m{=}5$, and slightly lower performance for $m{=}10$ and $m {\sim} U(0,10)$.

\subsection{Performance comparison for ATLAS dataset}
\renewcommand\arraystretch{1.3}
\begin{table*}[!t]
\caption{Comparison of dice coefficients among different methods for the ATLAS dataset when bounding boxes at different precision levels are considered.}
\centering
\setlength{\tabcolsep}{5pt}
\begin{tabular}{ccccc}
\hline
\hline
Methods & $m=0$ & $m=5$ & $m=10$ & $m \sim U(0,10)$ \\
\hline
FSIS & 0.512 (0.292) & NA & NA & NA \\
Deep cut \cite{rajchl2016deepcut} & 0.375 (0.246) & - & - & - \\
Global constraint \cite{kervadec2020bounding} & 0.474 (0.245) & - & - & - \\
MIL baseline & 0.408 (0.249) & 0.395 (0.240) & 0.357 (0.223) & 0.379 (0.237) \\
PA ($\alpha$-softmax) \cite{wang2021bounding} & 0.494 (0.236) & 0.451 (0.248) & 0.400 (0.254) & 0.456 (0.269) \\
PA ($\alpha$-quasimax) \cite{wang2021bounding} & 0.488 (0.240) & 0.448 (0.250) & 0.412 (0.241) & 0.437 (0.267) \\
PO ($\alpha$-softmax) & 0.463 (0.241) & 0.417 (0.225) & 0.373 (0.220) & 0.432 (0.215) \\
PO ($\alpha$-quasimax) & 0.470 (0.221) & 0.407 (0.241) & 0.381 (0.230) & 0.437 (0.218) \\
Proposed approach ($\alpha$-softmax) & 0.491 (0.233) & 0.464 (0.251) & 0.418 (0.246) & 0.487 (0.246) \\
Proposed approach ($\alpha$-quasimax) & 0.503 (0.245) & 0.462 (0.265) & 0.409 (0.236) & 0.464 (0.267) \\
\hline
\hline
\end{tabular}
\label{table:atlas_perf_comparison}
\end{table*}

Table \ref{table:atlas_perf_comparison} reports dice coefficients of the proposed approach supervised by bounding boxes at different precision levels for the ATLAS dataset. As can be noted, the proposed approach gets decreased performance for $m{=}5$, $m{=}10$, and $m {\sim} U(0,10)$ when compared with $m{=}0$. In particular, for $m{=}10$, the proposed approach has dice coefficient of 0.418 (a 14.87\% reduction in performance) when using $\alpha$-softmax function and 0.409 (a 18.69\% reduction) when employing $\alpha$-quasimax function. These results show that severe or very severe errors in bounding boxes could decrease the segmentation performance greatly for the proposed approach.

For comparison, Table \ref{table:atlas_perf_comparison} also shows results of the MIL baseline. It gets much lower dice coefficients when compared with the proposed approach at all precision levels. 

In Table \ref{table:atlas_perf_comparison}, we also report the results of PA and PA approaches. Both approaches get much lower dice coefficients when compared with the proposed approach at different precision levels. These results certify the effectiveness of both parallel transformation based MIL and polar transformation based MIL in the proposed approach.

Moreover, Table \ref{table:atlas_perf_comparison} also lists the results of Deep cut and Global constraint approaches reported in study \cite{kervadec2020bounding}, where only the results for $m{=}0$ are available. The dice coefficients of these two methods are much lower than those of the proposed approach.

Finally, FSIS achieves dice coefficient of 0.512, which is close to the results of the proposed approach for $m{=}0$, and much higher for $m{=}5$, $m{=}10$, and $m {\sim} U(0,10)$.

\subsection{Visualization of the origin in the polar transformation}
\begin{figure}[!t] 
	\centering
	\setlength{\tabcolsep}{2pt}
	\begin{tabular}{ccc}
	\includegraphics[trim=0in 0in 0in 0in,clip,width=1.1in]{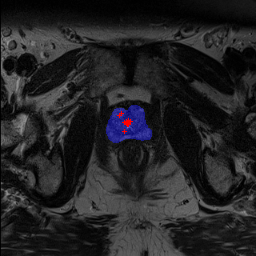} &
	\includegraphics[trim=0in 0in 0in 0in,clip,width=1.1in]{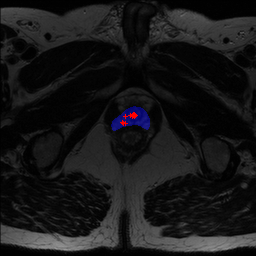} &
	\includegraphics[trim=0in 0in 0in 0in,clip,width=1.1in]{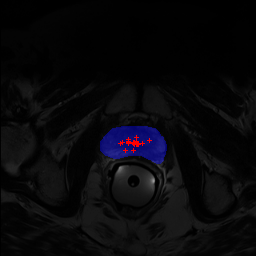} \\ 
	\multicolumn{3}{c}{(a) PROMISE12, $m=0$} \\
	\includegraphics[trim=0in 0in 0in 0in,clip,width=1.1in]{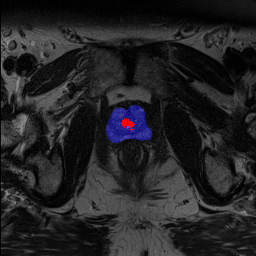} &
	\includegraphics[trim=0in 0in 0in 0in,clip,width=1.1in]{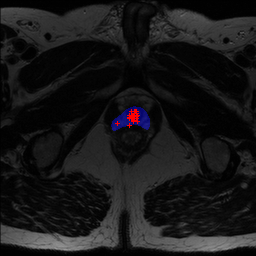} &
	\includegraphics[trim=0in 0in 0in 0in,clip,width=1.1in]{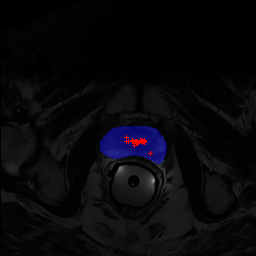} \\ 
	\multicolumn{3}{c}{(b) PROMISE12, $m \sim U(0,10)$} \\
	\includegraphics[trim=0in 0in 0in 0in,clip,width=1.1in]{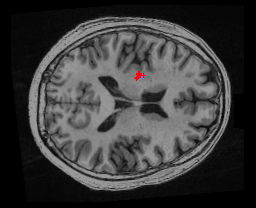} &
	\includegraphics[trim=0in 0in 0in 0in,clip,width=1.1in]{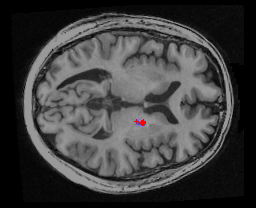} &
	\includegraphics[trim=0in 0in 0in 0in,clip,width=1.1in]{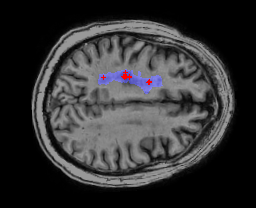} \\ 
	\multicolumn{3}{c}{(c) ATLAS, $m=0$} \\
	\includegraphics[trim=0in 0in 0in 0in,clip,width=1.1in]{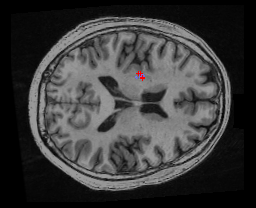} &
	\includegraphics[trim=0in 0in 0in 0in,clip,width=1.1in]{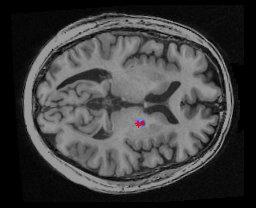} &
	\includegraphics[trim=0in 0in 0in 0in,clip,width=1.1in]{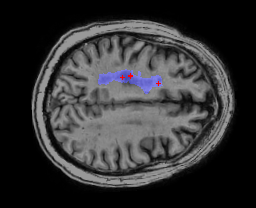} \\ 
	\multicolumn{3}{c}{(d) ATLAS, $m \sim U(0,10)$} \\
	\end{tabular}
	\caption{Selected origins in the polar transformation, where the selected origins are denoted by red plus signs and the pixel-wise ground truths of segmentation are marked by blue color.}
	\label{fig:polar_origin}
\end{figure}

In the proposed approach, the origin $O$ in the polar transformation was determined during training. To verify that the such automatic selection process indeed yields valid origin $O$ (i.e. it is located inside the object in the bounding box), we show the selected origins of several example images in Figure \ref{fig:polar_origin}. In these plots, the models of the proposed approach with $\alpha$-softmax function obtained at the end of each epoch are considered for origin selection. The selected origins are marked by red plus signs and the pixel-wise ground truths of segmentation are indicated by blue color. In Figure \ref{fig:polar_origin}, bounding boxes at two different precision levels are considered: $m{=}0$ denoting the use of accurate bounding boxes and $m {\sim} U(0,10)$ indicating the use of simulated real bounding box annotations. From Figure \ref{fig:polar_origin}, all selected origins are located in the object, verifying that the proposed approach is able to select origins correctly during training. 

\section{Conclusion}
This study investigates whether it is possible to maintain good segmentation performance for loose bounding box supervision. Extending the previous parallel transformation based MIL, it developed an MIL strategy based on polar transformation to assist image segmentation. Moreover, a weighted smooth maximum approximation was introduced to incorporate the observation that pixels closer to the origin of the polar transformation are more likely to belong to the object in the bounding box. In the experiments, the proposed approach was evaluated on two public datasets using dice coefficient for bounding boxes at different precision levels. The results demonstrate the superior performance of the proposed approach for bounding boxes at different precision levels and the robustness of the proposed approach for bounding boxes with mild and moderate errors.

\bibliographystyle{unsrt}
\bibliography{reference}  %%% Uncomment this line and comment out the ``thebibliography'' section below to use the external .bib file (using bibtex) .





% \end{thebibliography}

\end{document}